\documentclass{article}
\usepackage{times}
\usepackage{latexsym}
\usepackage[T1]{fontenc}
\usepackage[utf8]{inputenc}
\usepackage{spconf,amsmath,graphicx}
\usepackage{microtype}
\usepackage{multirow}

\usepackage{inconsolata}
\usepackage{color, colortbl}
\usepackage{xcolor}
\usepackage{enumitem}
\definecolor{White}{gray}{1.0}
\definecolor{ashgrey}{rgb}{0.94, 0.92, 0.84}
\usepackage{graphicx}
\RequirePackage{graphicx}
\RequirePackage{amssymb,amsmath,bm}
\RequirePackage{textcomp}
\RequirePackage{booktabs}
\RequirePackage[textfont=it,tableposition=top]{caption}

\pdfoutput=1
\title{Multilingual Word Error Rate Estimation: e-WER3}
\name{Shammur Absar Chowdhury, Ahmed Ali}
\address{Qatar Computing Research Institute, HBKU, Doha, Qatar}
\begin{document}
\ninept
\maketitle
\begin{abstract}

The success of the multilingual automatic speech recognition systems empowered many voice-driven applications. However, measuring the performance of such systems remains a major challenge, due to its dependency on manually transcribed speech data in both mono- and multilingual scenarios. In this paper, we propose a novel multilingual framework -- eWER3 -- jointly trained on acoustic and lexical representation to estimate word error rate. We demonstrate the effectiveness of eWER3 to \textit{(i)} predict WER without using any internal states from the ASR and \textit{(ii)} use the multilingual shared latent space to push the performance of the close-related languages.
We show our proposed multilingual model outperforms the previous monolingual word error rate estimation method (eWER2) by an absolute 9\% increase in Pearson correlation coefficient (PCC), 
with better overall estimation between the predicted and reference WER. 


\end{abstract}
\section{Introduction}
\label{sec:intro}
Recent years have witnessed a surge in both mono- and multilingual speech recognition performances, with accuracy comparable or even outperforming the human performance on established benchmarks \cite{xiong2016achieving,saon2017english}. 
With such success, automatic speech recognition (ASR) systems have been commoditized as speech processing pipelines in many voice-driven applications such as personal assistant devices and broadcast media monitoring among others. However, our means of evaluating the usefulness of the ASR output have remained largely unchanged. 

Word error rate (WER) is the standard measure for evaluating the performance of ASR systems. To obtain a reliable estimation of the WER, a minimum of two hours of manually transcribed test data is typically required -- a time-consuming and expensive process. 
Often voice-driven applications require quick quality estimation of the automated transcription, which is not feasible with such traditional reference-based measures. Moreover, even with offline applications, it is not always viable to obtain gold references (especially in multilingual scenarios) to evaluate the transcription quality.
Thus, there is a need to develop techniques that can automatically estimate the quality of the ASR transcription without such manual effort \cite{negri2014quality,kiseleva2016understanding} and handle multilingualism.

Several studies have explored the automatic estimation of the WER. These studies included a large set of extracted features (with/without internal access to the ASR system) to train neural regression or classification models \cite{jalalvand2016transcrater, ali-ewer, ali2020word}. Some studies proposed a novel neural zero-inflated model \cite{fan2019neural}, while others model uncertainty \cite{ vyas2019analyzing} in predictions to handle different challenges. However, all these studies are conducted with networks directly trained and tested in monolingual settings.

\begin{figure}[hbt!]
\begin{center}
\scalebox{0.5}{

\includegraphics[]{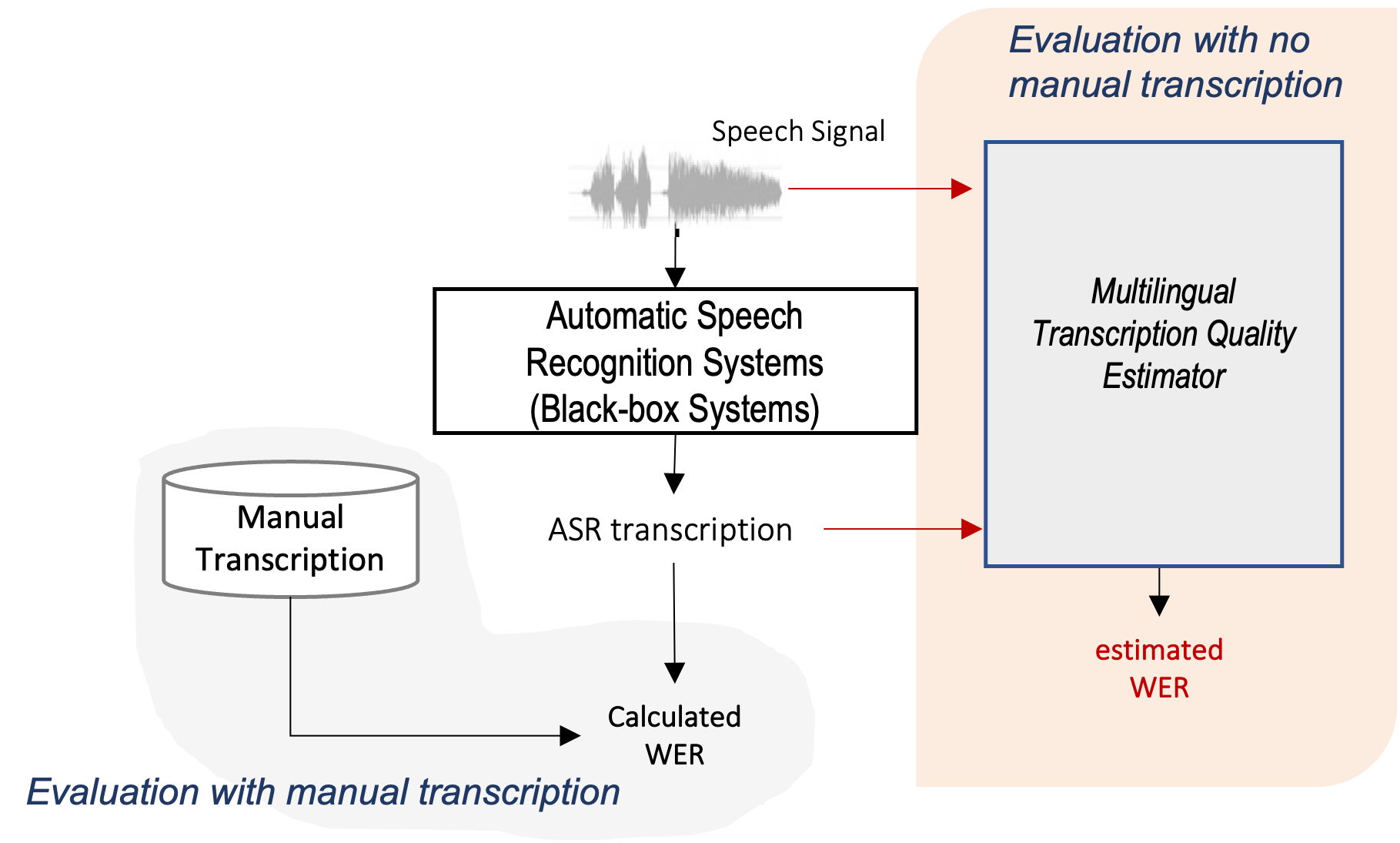}
}

\caption{Overview of the study concept and proposed framework.  } 
\vspace{-0.5cm}
\label{fig:overview} 
\end{center}
\end{figure}

In this work, we design a single multilingual end-to-end model capable of estimating WER given the raw audio and the automatic transcription from different (mono- and multilingual) off-the-shelf ASR systems without having access to the ASR's internal feature representation (the concept is shown in Figure \ref{fig:overview}). For this, we entail the large self-supervised pretrained models as feature extractor and exploits the available multilingual corpora.

We evaluate our results using \textit{Arabic}, \textit{Italian}, \textit{Spanish}, \textit{English} and \textit{Russian} -- test sets. We train a monolingual estimator and compare it with our proposed multilingual model to show its efficacy for better performance.
Our contributions are:
\begin{itemize}
\item Design the first multilingual WER estimation without using any internal features from the ASR (black-box);
\item Compare our method with previous state-of-the-art results (e-wer \cite{ali-ewer} and e-wer2 \cite{ali2020word});
\item Analyse the effect of imbalanced WER distribution on the estimator's performance and propose a new sampling technique.
\end{itemize}


\section{E2E Multilingual WER Estimator}
\begin{figure*}[hbt!]
\begin{center}
\scalebox{0.7}{
\includegraphics[]{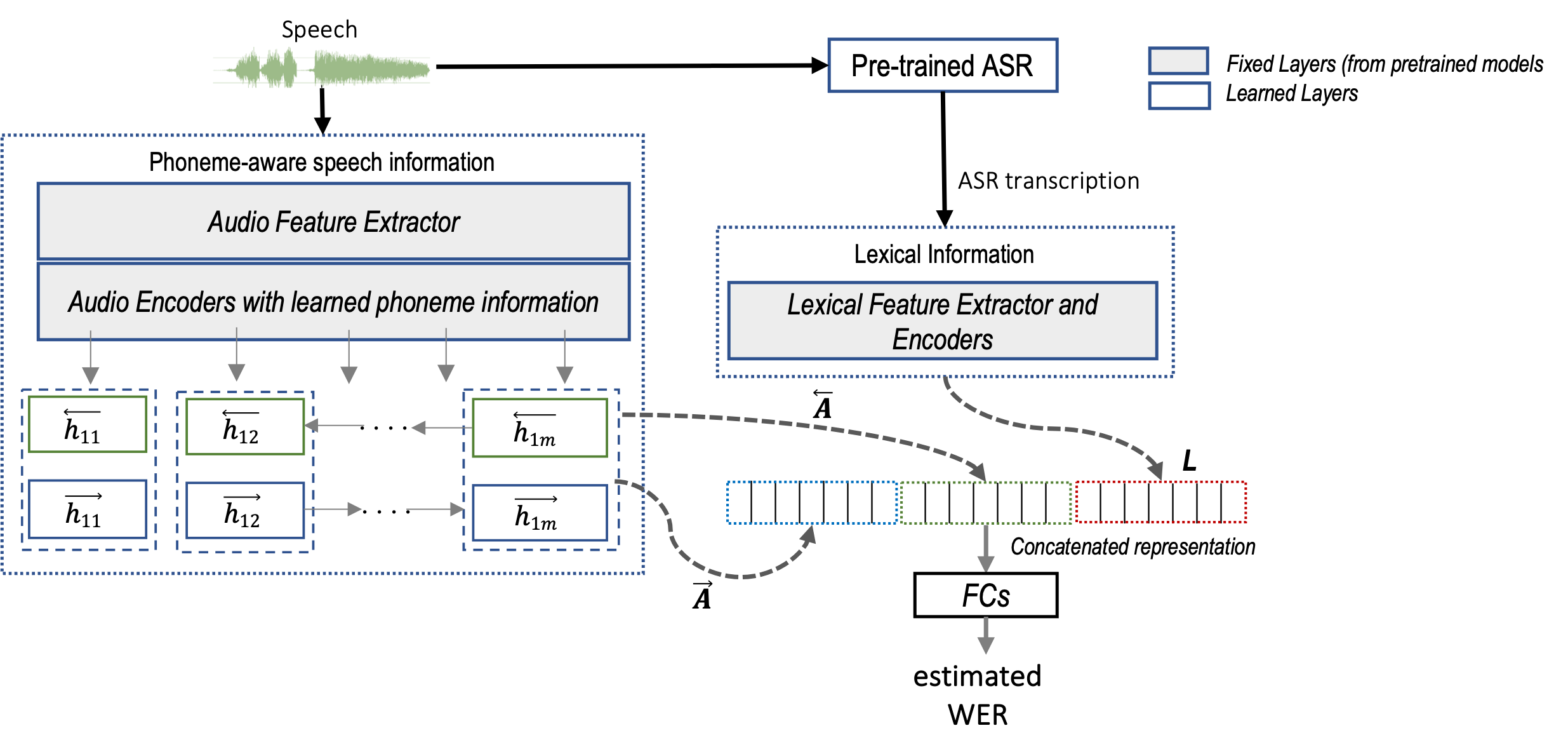}
}
\vspace{-0.3cm}
\caption{End-to-End Architecture used to estimate WER. $A^*$ : acoustic representation from BLSTM and L: the lexical representation.} 

\vspace{-0.3cm}
\label{fig:framework} 
\end{center}
\end{figure*}

Figure \ref{fig:framework} shows an overview of the end-to-end system architecture designed to estimate speech recognition WER with no gold-standard reference transcription. As input to the estimator, we first pass raw audio along with its automatic transcription obtained from the speech recognition systems. We extract the speech and lexical representations and utilize these representations jointly to train the multilingual regression model.

\begin{table*} [!ht]
\centering
\scalebox{0.98}{
\begin{tabular}{l||c|c|c||c|c} 
\multicolumn{1}{c||}{\textit{Lang}} & \textbf{ASR Trained on}                                                                     & \textbf{Architecture~} & \textbf{ASR Type} & \textbf{ Estimator Trained using} & \textbf{ Estimator Tested on} \\ 
\hline
English                            &  LibriSpeech ($960$hours)                                                                 & Conformer              & Mono              & LibriSpeech + TEDLIUM3 dev &     TEDLIUM3 test         \\ 
\hline
Spanish                             & CommonVoice (CV-ES) \cite{ardila2019common}                                                                         & Conformer              & Mono              & CV-ES dev & CV-ES test              \\ 
\hline
Italian                             & CommonVoice (CV-IT)                                                                         & Conformer              & Mono              & CV-IT dev & CV-IT test              \\ 
\hline
Russian                             & CommonVoice (CV-RU)                                                                        & Conformer              & Mono              & CV-RU dev & CV-RU test              \\ 
\hline
Arabic                               & \begin{tabular}[c]{@{}c@{}}QASR (Arabic) +\\ LibriSpeech (English) (200hrs) \cite{mubarak2021qasr} \\\end{tabular} & Conformer              & Multi            & MGB2 \cite{khurana2016qcri} + QASR dev & SUMMA \cite{magdy2012summarization}              \\ 
\hline
\end{tabular}
}
\vspace{+0.2cm}
\caption{Description of the ASR systems used to train and/or test the proposed estimator along with the Estimator training and test set. Lang. shows the language data used to train the estimator. }
\label{tab:asr_description}
\vspace{-0.4cm}
\end{table*}

\begin{table*}[!ht]
\centering
\scalebox{1.0}{
\begin{tabular}{l|c|c|c|c}
\multicolumn{1}{c|}{\textit{Train (Dev | Test)}} & \textbf{T.D} (hours) & \textbf{A.D} (secs) & \textbf{A.U} (words) & \textbf{\#} \\ \hline
Arabic & 10.36 (0.56 | 2.94) & 6.14 (6.08 | 7.52) & 12.66 (12.45 | 13.96) & 6077 (332 | 1410) \\ \hline
English & 3.72 (0.31 | 2.62) & 5.83 (5.65 | 8.18) & 16.68 (16.04 | 23.95) & 2295 (196 | 1151) \\ \hline
Spanish & 7.37 (0.84 | 17.31) & 6.05 (6.11 | 6.12) & 10.11 (10.17 | 9.84) & 4384 (496 | 10179) \\ \hline
Italian & 4.94 (0.53 | 12.26) & 5.94 (5.91 | 6.13) & 10.2 (10.13 | 9.74 ) & 2991 (322 | 7200) \\ \hline
Russian & 1.73 (0.17 | 9.57) & 5.94 (5.83 | 5.99) & 9.86 (9.72 | 9.58) & 1045 (102 | 5748) \\ \hline
\end{tabular}%
}
\vspace{+0.2cm}
\caption{Train, Dev and Test (Dev and Test in bracket seperated by |) Data Description. T.D: Total dataset duration in hours; A.D: Average utterance duration in seconds; A.U: Average utterance length in tokens; \#:Total instances.}
\label{tab:data_description}
\vspace{-0.9cm}
\end{table*}

\paragraph*{Acoustic representation:} We use XLSR-$53$ to extract phoneme aware speech representation. The XLSR-$53$ model is a multilingual variation of wav2vec $2.0$ model fine-tuned on cross-lingual phoneme-recognition task \cite{conneau2020unsupervised,xu2021simple}. For our study, we remove the output (language model head) and use the representation only. We use XLSR-$53$ as a feature extractor, which includes a cascaded temporal convolutional network to map raw audio, $X=\{x_1, x_2 .., x_n\}$ to the latent speech representation $Z=\{z_1, z_2 ..,z_t\}$. This latent information is then passed through 24 Transformer \cite{devlin2018bert} blocks with model dimension of $1,024$ and $16$ attention heads, to capture contextual representations, $C$ ($g : Z \mapsto C $). We  then pass the frame-wise representation to a bi-directional LSTM and extracted the last step representations ($\overleftarrow{A},\overrightarrow{A} $).


\paragraph*{Lexical representation:} Simultaneously, to extract the lexical embeddings, we pass the ASR transcription to the XLM-RoBERTa-Large model \cite{conneau2019unsupervised}, pretrained using 100 different languages. The pretrained model follows the same architecture as BERT \cite{devlin2018bert}, with $24$-layers of transformer modules -- with $1,024$ hidden-state, and $16$ attention heads. The model uses a byte-level BPE as a tokenizer and outputs the sequence of hidden-states for the whole input sentence. To obtain the final lexical representation ($L$), we averaged the embeddings over the sequences. 

\paragraph*{Combining representations:} We concatenate the output representations from the acoustic and lexical module ($\overleftarrow{A}+\overrightarrow{A}+L$) and pass it through two fully-connected layers, before the output layer, for the regression task.


\section{Experimental Setup}

\subsection{Speech Recognition Systems}
To train the estimator, we opt for the current state-of-the-art conformer \cite{conformer} based end-to-end speech recognition systems (see Table \ref{tab:asr_description}). 

For the Spanish, Italian, and Russian ASR systems, the models are trained using their respective CV train sets. The model has $12$ encoder layers and $6$ decoder layers each with $2,048$ encoder/decoder units from FFN and $4$ attention heads with $256$ transformation dimensions and $15$ CNN kernels. 

As for the English ASR, we use a large conformer model with $12$ encoder and $6$ decoder layers containing $8$ attention heads with $512$ transformation dimensions and $31$ CNN kernels.
This large ASR is trained using the well-known $960$ hours of librispeech data. 
We use similar architecture for multilingual Arabic ASR \cite{chowdhury2021towards} trained with Arabic QASR \cite{mubarak2021qasr} along with English librispeech data. 


\begin{figure}[hbt!]
\centering
\scalebox{0.55}{
\includegraphics[]{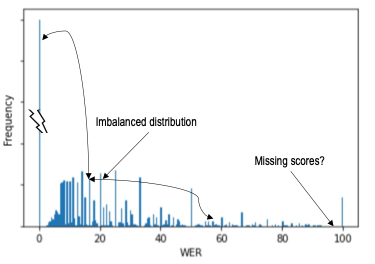}
}
\caption{Data imbalance and missing target values.}
\vspace{-0.3cm}
\label{fig:data-imbalance} 
\end{figure}
\subsection{Data} %

\paragraph*{Data Preparation:} We train the multilingual estimator using the dataset mentioned in Table \ref{tab:asr_description}.
The input audio to the estimator is first down-sampled to 16KHz to match the ASR input sample rate. For training the model, we select audio instances with a duration of 10 seconds or less; this is based on the upper tail of overall duration distribution.

\paragraph*{Imbalanced Distribution:} Given the remarkable performance of the current end-to-end ASR models, the  WER often exhibits imbalanced distributions, where certain target values have significantly fewer observations than others.  
In this case, the majority of the training set has a WER of `$0$', making the training data highly skewed (see Figure \ref{fig:data-imbalance}). Moreover, the dataset shows a tendency of missing data for certain target values, thus making the task more challenging. 
In order to handle the abundance of `$0$' WER scores in the training set, we sampled $n$ instances from each language with $WER=0$. We determine the value $n$ based on the sum of instances falls under the next two most frequent score groups.

\paragraph*{Data Split:} For our dev set, we divide the training dataset into 10 bins of target WER, with equal intervals such as $[0,10), [10,20) \cdots$ $[90,100]$. From each bin, we then randomly sample $\approx$10\% of the instances to create the validation set.
The details of the resultant (balanced) split are shown in Table \ref{tab:data_description}.

\paragraph*{Estimator Output:} As the output score of the estimator, it is worth noting that we bound the target value (WER) of the estimator to a range of [0,1] (i.e. 0 - 100\%).\footnote{If WER $>100\%$, the value is scaled down to 100.}

\begin{table*}[!ht]
\centering
\scalebox{1.0}{
\begin{tabular}{l|c|c|c}
\multicolumn{1}{c|}{\textit{Ar:SUMMA}} & \textbf{PCC} & \textbf{RMSE} & \textbf{Input to the Estimator} \\ \hline
eWER & 0.66 & 0.35 & Lexical + Grapheme + Decoder + Numerical \cite{ali-ewer}  \\ \hline
eWER2 & 0.72 & 0.20 & MFCC+ Lexical + Phonetic \cite{ali2020word} \\ \hline
\textbf{\begin{tabular}[c]{@{}l@{}}eWER3 mono\end{tabular}} & \textbf{0.75} & \textbf{0.14} & Raw Audio, Lexical Transcription \\ \hline
\end{tabular}%
}
\vspace{+0.2cm}
\caption{Monolingual (Arabic) transcription quality estimator results on Ar:SUMMA test set. }
\vspace{-0.4cm}
\label{tab:mono-result}
\end{table*}

\begin{table}[!htb]

\centering
\scalebox{1.0}{

\begin{tabular}{l|c|c|c|c|c}
\multicolumn{1}{c|}{Sets} & PCC & RMSE & eWER3 & WER &  \# \\ \hline\hline
\rowcolor{White}
\multicolumn{6}{c}{Monolingual Estimator Model - Arabic} \\ \hline\hline
\multicolumn{1}{l|}{Ar:SUMMA} & \multicolumn{1}{c|}{0.75} & \multicolumn{1}{c|}{0.14} & \multicolumn{1}{c|}{16.0\%} & \multicolumn{1}{c|}{18.0\%}  & 1410 \\
\rowcolor{ashgrey}
\multicolumn{1}{l|}{It:CV} & \multicolumn{1}{c|}{0.45} & \multicolumn{1}{c|}{0.32} & \multicolumn{1}{c|}{41.0\%} & \multicolumn{1}{c|}{17.0\%}  & 7200 \\
\hline\hline
\rowcolor{White}
\multicolumn{6}{c}{Monolingual Estimator Model - English} \\ \hline\hline
\multicolumn{1}{l|}{En:TedL} & \multicolumn{1}{c|}{0.62} & \multicolumn{1}{c|}{0.14} & \multicolumn{1}{c|}{7.0\%} & \multicolumn{1}{c|}{12.0\%} &  1151 \\
\rowcolor{ashgrey}
\multicolumn{1}{l|}{It:CV} & \multicolumn{1}{c|}{0.49} & \multicolumn{1}{c|}{0.19} & \multicolumn{1}{c|}{10.0\%} & \multicolumn{1}{c|}{17.0\%} &  7200 \\
\hline\hline

\rowcolor{White}
\multicolumn{6}{c}{Multilingual Estimator Model} \\ \hline\hline
\multicolumn{1}{l|}{Ar:SUMMA} & \multicolumn{1}{c|}{0.74} & \multicolumn{1}{c|}{0.15} & \multicolumn{1}{c|}{15.0\%} & \multicolumn{1}{c|}{18.0\%} & 1410 \\
\multicolumn{1}{l|}{It:CV} & \multicolumn{1}{c|}{0.60} & \multicolumn{1}{c|}{0.17} & \multicolumn{1}{c|}{14.0\%} & \multicolumn{1}{c|}{17.0\%}  & 7200 \\
\multicolumn{1}{l|}{Es:CV} & \multicolumn{1}{c|}{0.53} & \multicolumn{1}{c|}{0.14} & \multicolumn{1}{c|}{13.0\%} & \multicolumn{1}{c|}{11.0\%} &  10179 \\
\multicolumn{1}{l|}{En:TedL} & \multicolumn{1}{c|}{0.66} & \multicolumn{1}{c|}{0.14} & \multicolumn{1}{c|}{8.0\%} & \multicolumn{1}{c|}{12.0\%} &  1151 \\
\multicolumn{1}{l|}{Ru:CV} & \multicolumn{1}{c|}{0.51} & \multicolumn{1}{c|}{0.12} & \multicolumn{1}{c|}{6.0\%} & \multicolumn{1}{c|}{7.0\%} &  5748 \\
\hline
\end{tabular}
}
\vspace{+0.2cm}
\caption{Reported performance of monolingual and multilingual WER estimator on Arabic (Ar), English (En), Italian (It), Spanish (Es) and Russian (Ru) test sets.}
\label{tab:results}

\vspace{-0.3cm}
\end{table}

\subsection{WER Estimator Design}
\noindent\textbf{Model Parameters}: We train the end-to-end WER estimator using a an Adam optimizer for $20$ epochs with a learning rate of $1e-3$ and a dropout-rate of $0.1$, and freeze the parameters of the pretrained self-supervised models. In the acoustic pipeline, we use one layer of BiLSTM model and for the joint training, we opt for two fully-connected layers ($600$, $32$ neurons) with ReLU activation function. As for the loss function, we use mean squared error. Same architecture and hyperparameters are used to train mono- and multilingual models with balanced and natural distribution data.


\subsection{Evaluation Measures}
Given the uneven scores distribution (towards small WER value), we use Pearson correlation coefficient (PCC) as our main evaluation metric. However, we also report root mean square error (RMSE) to compare with previous studies \cite{ali-ewer, ali2020word}.  
Moreover, to effectively estimate the eWER3 for the complete test set, we report weighted WER:  ($eWER3 = \frac{\sum \widehat{WER_{utt}}*Dur(utt)}{\sum^n Dur}$) using the utterance level estimated WER ($\widehat{WER_{utt}}$) and the corresponding duration ($Dur(utt)$).



\section{Results and Discussion}
\subsection{Monolingual Comparison}
We benchmark the proposed framework eWER3 in a monolingual setting (Arabic) and compare it with the previous estimation models -- eWER and eWER2. 
The results, reported in Table \ref{tab:mono-result}, show that our model outperforms both eWER and eWER2 with an absolute increase of $3$\% and $9$\% in PCC, and a decrease of $21$\% and $6$\% in RMSE respectively. Such improvement indicates the estimation power of our architecture without using any additional feature from the ASR decoder.

Moreover, when the monolingual models (for both Arabic and English -- in Table \ref{tab:results}) were tested in the cross-lingual Italian dataset, both models' performance (both in correlation coefficient and RMSE) decrease drastically. Yet, it is observed that the Italian test set benefits more from the English monolingual model with RMSE: $0.19$ compared to RMSE:$0.32$ in the Arabic model. 
Thus indicating the potential advantage of having shared latent space for close languages in multilingual settings.



\begin{figure}[hbt!]
\begin{center}
\scalebox{0.6}{

\includegraphics[]{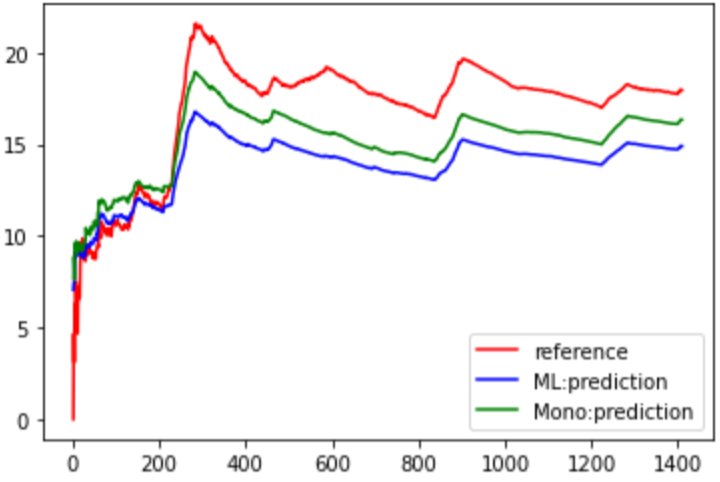}
}
\caption{Cumulative WER over time with all sentences for (a) Arabic SUMMA test set. X-axis is the number of instances and Y-axis is Aggregated WER in \%. }

\label{fig:ar-aggwer} 
\end{center}
\vspace{-0.65cm}
\end{figure}

\begin{figure}[hbt!]
\centering
\scalebox{0.6}{
\includegraphics[]{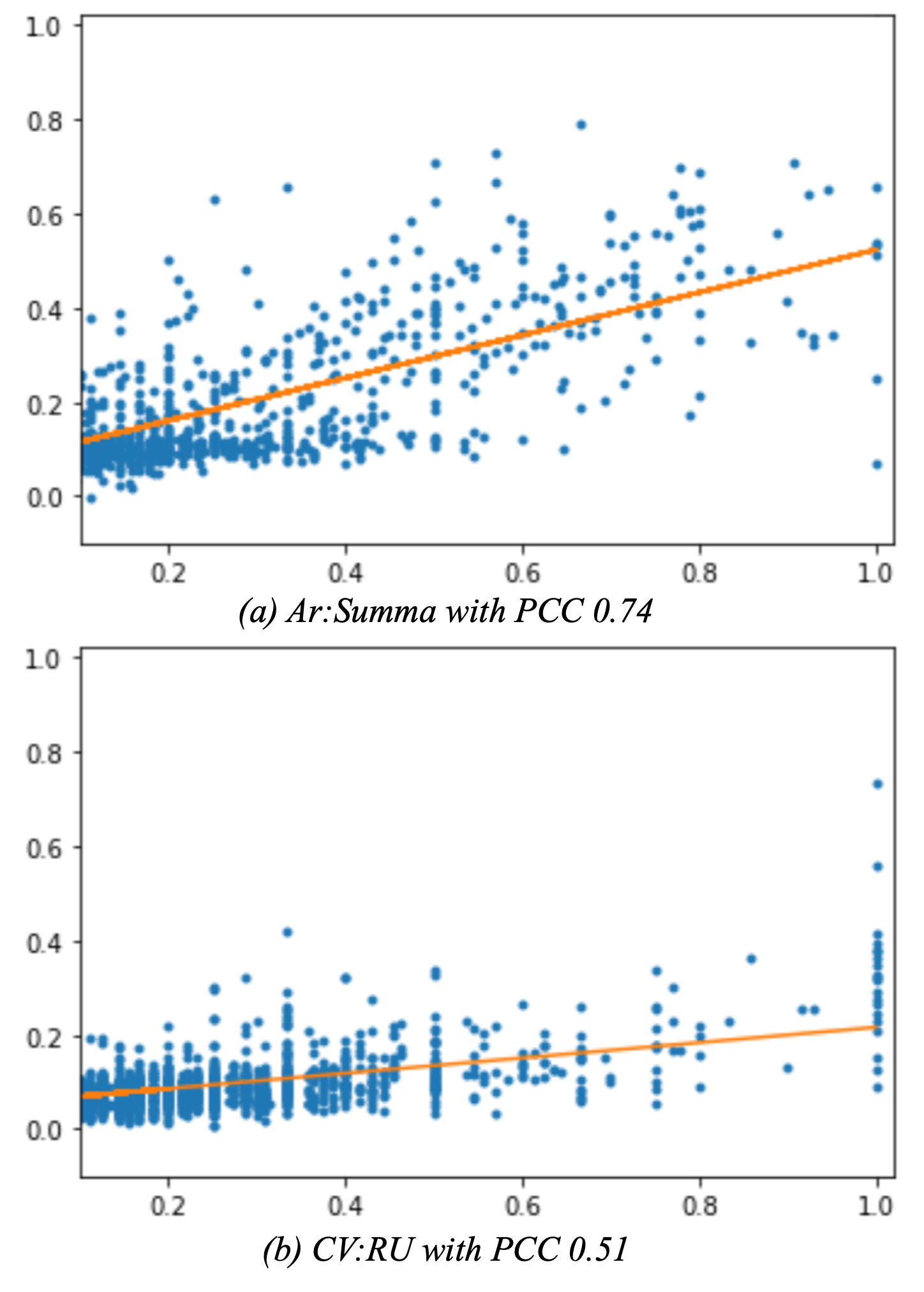}
}
\caption{Scatter Plot for test sets with highest PCC 0.74 -- Arabic (a); and the lowest PCC 0.51 -- Russian.}
\label{fig:scatter} 
\end{figure}

\subsection{Multilingual E2E Estimator Performance}


Table \ref{tab:results} shows that the multilingual model gives a comparable correlation and RMSE compared to the monolingual models. We notice the 4\% performance increase in PCC for English test set when in multilingual setting, showing the added advantage of such a multilingual estimator.
Furthermore, for all the language test sets (Arabic, English, Italian, Spanish, Russian), in addition to a smaller RMSE and significant correlation -- per utterance, the overall predicted WER  is also within a close range ($0-5$ points) of the actual WER. 

For brevity, we present the Arabic Summa test set's cumulative WER aggregated over the sentences in Figure \ref{fig:ar-aggwer} and the corresponding scatter plot for the Arabic (best PCC obtained) and Russian (lowest performing PCC) test sets in Figure \ref{fig:scatter}. 


\subsection{Effects of Imbalanced Data and Sampling}
\label{ssec:imb}
We analyse the effect of training the model with sampled data (Model Sampled: $\psi$) instead of natural distribution (Model Natural:$\varphi$). With respect to the $\psi$, we noticed $\varphi$ has a slightly better correlation coefficient, yet has higher RMSE-values and large difference in aggregated estimated eWER3 than the Oracle WER. For example, for ES:CV test set, $\varphi(PCC)=0.58$, $\varphi(RMSE)=0.15$, $\varphi(eWER3)=7.0\%$, whereas, $\psi(PCC)=0.53$, $\psi(RMSE)=0.14$, $\psi(eWER3)=13.0\%$.\footnote{A higher RMSE and overall WER difference is seen for other datasets while using natural distribution. }

\begin{figure}[hbt!]
\begin{center}
\scalebox{0.55}{

\includegraphics[]{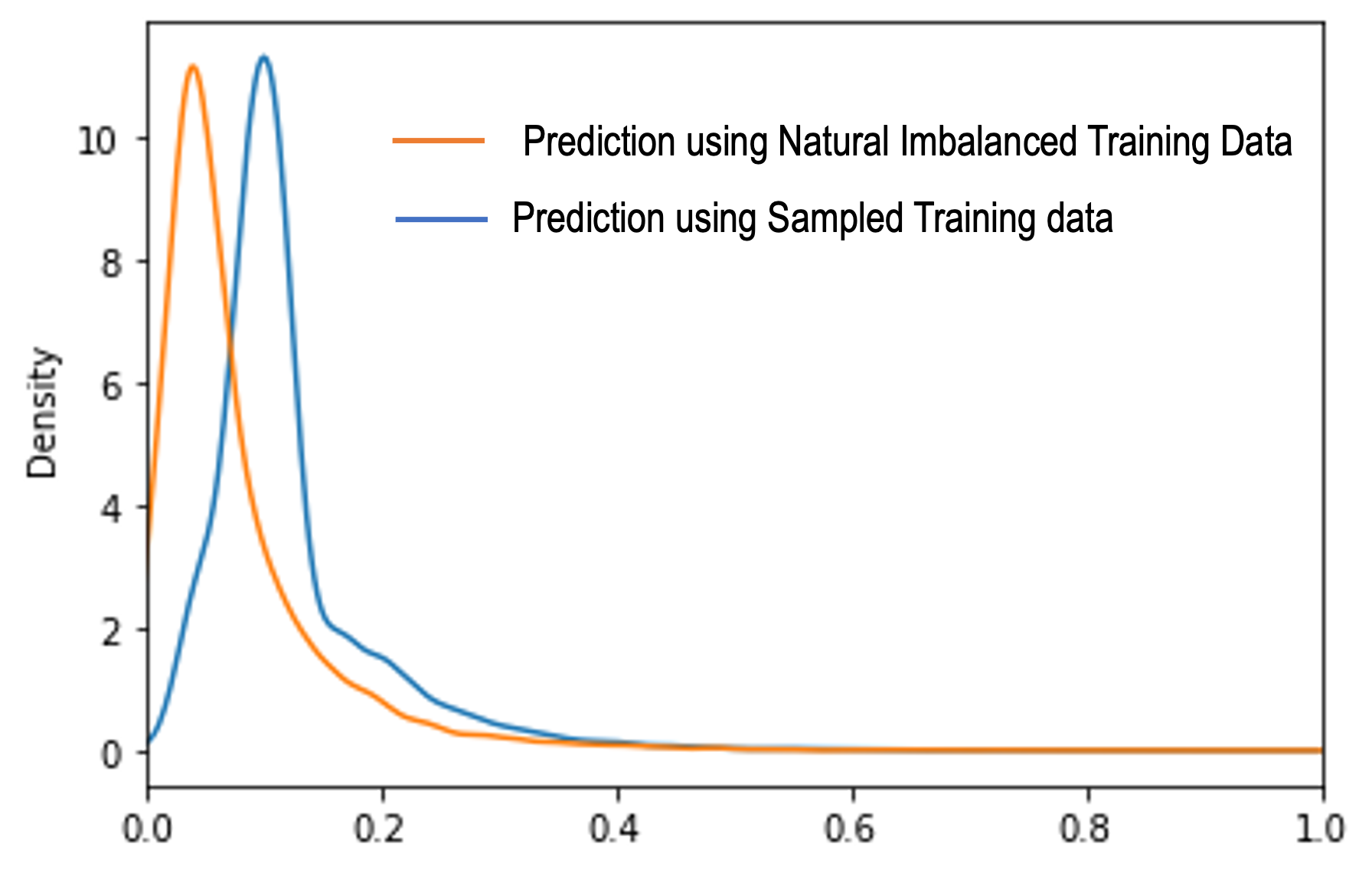}
}
\caption{Density curves using estimated WER for the multilingual model trained using sampled distribution (blue line) and natural distribution (orange) train set, showing the effect of imbalanced data labels. x-axis represents WER. The prediction is from aggregated in-language test sets.}

\label{fig:density} 
\end{center}
\vspace{-0.6cm}
\end{figure}

The density curve, from $\varphi$ and $\psi$ model predictions (Figure \ref{fig:density}), indicates that with natural distribution the model ($\varphi$) learns to predict lower WER better than the $\psi$. However, the prediction is scaled down to a lower range (see the shift in the peak of both the curves) thus increasing RMSE and the difference between the overall predicted eWER3 and Oracle WER. This is a potential limitation of the current study and a future endeavor for experimenting with zero-inflated output layers \cite{fan2020neural} for such a multilingual network. 

\section{Conclusion}

In this study, we propose a novel framework, for estimating multilingual WER without the need of manual transcription. Our proposed framework is a joint acoustic-lexical model exploiting the self-supervised learning paradigm. Using a small subset of languages, our results suggest the efficacy of such model to predict utterance-level and overall WER for the test sets. When compared with monolingual models, the multilingual framework performs comparably for the distant languages (e.g., Arabic) while boosting the performance of the close languages (e.g., $En_{mono}$: 0.62 PCC {\em vs} $En_{multi}$: 0.66 PCC). 
The current study can be used as a proof of concept to utilize the representation models to design such a predictor for an ASR. We exploit pretrained models as a feature-extractor for computational feasibility. 
In the future, we will focus on improving the performance by handling the imbalanced target distribution, with improved neural architecture and cover more languages.

\bibliographystyle{IEEEbib}
\bibliography{main}

\begin{thebibliography}{10}

\bibitem{xiong2016achieving}
Wayne Xiong, Jasha Droppo, Xuedong Huang, Frank Seide, Mike Seltzer, Andreas
  Stolcke, Dong Yu, and Geoffrey Zweig,
\newblock ``Achieving human parity in conversational speech recognition,''
\newblock {\em arXiv preprint arXiv:1610.05256}, 2016.

\bibitem{saon2017english}
George Saon, Gakuto Kurata, Tom Sercu, Kartik Audhkhasi, Samuel Thomas,
  Dimitrios Dimitriadis, Xiaodong Cui, Bhuvana Ramabhadran, Michael Picheny,
  Lynn-Li Lim, Bergul Roomi, and Phil Hall,
\newblock ``English conversational telephone speech recognition by humans and
  machines,''
\newblock {\em arXiv preprint arXiv:1703.02136}, 2017.

\bibitem{negri2014quality}
Matteo Negri, Marco Turchi, Jos{\'e}~GC de~Souza, and Daniele Falavigna,
\newblock ``Quality estimation for automatic speech recognition,''
\newblock in {\em Proceedings of COLING 2014, the 25th International Conference
  on Computational Linguistics: Technical Papers}, 2014, pp. 1813--1823.

\bibitem{kiseleva2016understanding}
Julia Kiseleva, Kyle Williams, Jiepu Jiang, Ahmed Hassan~Awadallah, Aidan~C
  Crook, Imed Zitouni, and Tasos Anastasakos,
\newblock ``Understanding user satisfaction with intelligent assistants,''
\newblock in {\em Proceedings of the 2016 ACM on Conference on Human
  Information Interaction and Retrieval}, 2016, pp. 121--130.

\bibitem{jalalvand2016transcrater}
Shahab Jalalvand, Matteo Negri, Marco Turchi, Jos{\'e}~GC de~Souza, Daniele
  Falavigna, and Mohammed~RH Qwaider,
\newblock ``Transcrater: a tool for automatic speech recognition quality
  estimation,''
\newblock {\em AC}, 2016.

\bibitem{ali-ewer}
Ahmed Ali and Steve Renals,
\newblock ``Word error rate estimation for speech recognition: e-{WER},''
\newblock in {\em ACL}, 2018.

\bibitem{ali2020word}
Ahmed Ali and Steve Renals,
\newblock ``Word error rate estimation without asr output: e-wer2,''
\newblock {\em Proc. Interspeech 2020}, pp. 616--620, 2020.

\bibitem{fan2019neural}
Kai Fan, Jiayi Wang, Bo~Li, Boxing Chen, and Niyu Ge,
\newblock ``Neural zero-inflated quality estimation model for automatic speech
  recognition system,''
\newblock {\em arXiv preprint arXiv:1910.01289}, 2019.

\bibitem{vyas2019analyzing}
Apoorv Vyas, Pranay Dighe, Sibo Tong, and Herv{\'e} Bourlard,
\newblock ``Analyzing uncertainties in speech recognition using dropout,''
\newblock in {\em ICASSP}, 2019.

\bibitem{ardila2019common}
Rosana Ardila, Megan Branson, Kelly Davis, Michael Henretty, Michael Kohler,
  Josh Meyer, Reuben Morais, Lindsay Saunders, Francis~M Tyers, and Gregor
  Weber,
\newblock ``Common voice: A massively-multilingual speech corpus,''
\newblock {\em arXiv preprint arXiv:1912.06670}, 2019.

\bibitem{mubarak2021qasr}
Hamdy Mubarak, Amir Hussein, Shammur~Absar Chowdhury, and Ahmed Ali,
\newblock ``{QASR}: {QCRI} {A}ljazeera speech resource a large scale annotated
  arabic speech corpus,''
\newblock in {\em ACL/IJCNLP (1)}, 2021.

\bibitem{khurana2016qcri}
Sameer Khurana and Ahmed Ali,
\newblock ``{QCRI} advanced transcription system ({QATS}) for the {Arabic
  Multi-Dialect Broadcast Media Recognition: MGB-2 Challenge},''
\newblock in {\em SLT}, 2016.

\bibitem{magdy2012summarization}
Walid Magdy, Ahmed Ali, and Kareem Darwish,
\newblock ``A summarization tool for time-sensitive social media,''
\newblock in {\em Proceedings of the 21st ACM international conference on
  Information and knowledge management}. ACM, 2012, pp. 2695--2697.

\bibitem{conneau2020unsupervised}
Alexis Conneau, Alexei Baevski, Ronan Collobert, Abdelrahman Mohamed, and
  Michael Auli,
\newblock ``Unsupervised cross-lingual representation learning for speech
  recognition,''
\newblock {\em arXiv preprint arXiv:2006.13979}, 2020.

\bibitem{xu2021simple}
Qiantong Xu, Alexei Baevski, and Michael Auli,
\newblock ``Simple and effective zero-shot cross-lingual phoneme recognition,''
\newblock {\em arXiv preprint arXiv:2109.11680}, 2021.

\bibitem{devlin2018bert}
Jacob Devlin, Ming-Wei Chang, Kenton Lee, and Kristina Toutanova,
\newblock ``Bert: Pre-training of deep bidirectional transformers for language
  understanding,''
\newblock {\em arXiv preprint arXiv:1810.04805}, 2018.

\bibitem{conneau2019unsupervised}
Alexis Conneau, Kartikay Khandelwal, Naman Goyal, Vishrav Chaudhary, Guillaume
  Wenzek, Francisco Guzm{\'a}n, Edouard Grave, Myle Ott, Luke Zettlemoyer, and
  Veselin Stoyanov,
\newblock ``Unsupervised cross-lingual representation learning at scale,''
\newblock {\em arXiv preprint arXiv:1911.02116}, 2019.

\bibitem{conformer}
Anmol Gulati, James Qin, Chung-Cheng Chiu, Niki Parmar, Yu~Zhang, Jiahui Yu,
  Wei Han, Shibo Wang, Zhengdong Zhang, Yonghui Wu, et~al.,
\newblock ``Conformer: Convolution-augmented transformer for speech
  recognition,''
\newblock {\em arXiv preprint arXiv:2005.08100}, 2020.

\bibitem{chowdhury2021towards}
Shammur~Absar Chowdhury, Amir Hussein, Ahmed Abdelali, and Ahmed Ali,
\newblock ``Towards one model to rule all: Multilingual strategy for dialectal
  code-switching {A}rabic {ASR},''
\newblock {\em Interspeech}, 2021.

\bibitem{fan2020neural}
Kai Fan, Bo~Li, Jiayi Wang, Shiliang Zhang, Boxing Chen, Niyu Ge, and Zhijie
  Yan,
\newblock ``Neural zero-inflated quality estimation model for automatic speech
  recognition system,''
\newblock in {\em Interspeech}, 2020.

\end{thebibliography}

\end{document}